\pdfoutput=1

\documentclass[11pt]{article}

\usepackage{acl}

\usepackage{times}
\usepackage{latexsym}

\usepackage[T1]{fontenc}

\usepackage[utf8]{inputenc}

\usepackage{microtype}

\input{macros.sty}

\usepackage{amsmath,amsfonts,bm}

\def\eqref#1{equation~\ref{#1}}

\def\1{\bm{1}}

\def\rvy{{\mathbf{y}}}

\DeclareMathAlphabet{\mathsfit}{\encodingdefault}{\sfdefault}{m}{sl}
\SetMathAlphabet{\mathsfit}{bold}{\encodingdefault}{\sfdefault}{bx}{n}

\definecolor{applegreen}{rgb}{0.13, 0.58, 0.32}
\definecolor{red}{rgb}{0.9, 0, 0}

\usepackage{todonotes}
\usepackage{booktabs}
\usepackage{multirow}
\usepackage{amsmath,amssymb}
\usepackage{algorithm, algpseudocode}
\usepackage{cleveref}
\usepackage[usestackEOL]{stackengine}
\usepackage[]{xcolor}
\usepackage{pdfpages}

\makeatletter
\def\dual#1{\expandafter\dual@aux#1\@nil}
\def\dual@aux#1/#2\@nil{\begin{tabular}{@{}c@{}}#1\\#2\end{tabular}}
\makeatother

\title{TempLM: Distilling Language Models into Template-Based Generators}

\author{Tianyi Zhang, Mina Lee${}^{*}$, Lisa Li${}^{*}$, Ende Shen${}^{*}$, Tatsunori B. Hashimoto \\
  Computer Science Department, Stanford University \\
  \texttt{\{tz58, minalee, xlisali, endeshen, thashim\}@stanford.edu}\\
}

\begin{document}
\maketitle
\begin{abstract}
  While pretrained language models (\lm{}s) have greatly improved text generation, they have also been known to produce unfaithful or inappropriate content.
  In contrast, classic template-based systems provide strong guarantees of faithfulness at the cost of fluency.
We propose \templm{}, which achieves the best of both worlds by distilling a \lm{} into a template-based generator.
On the E2E and SynthBio data-to-text datasets, we show that \templm{} is more faithful than the original \lm{} and is more fluent than prior template systems.
Notably, on an out-of-domain evaluation, \templm{} reduces a finetuned BART model's unfaithfulness rate from $83\%$ to $0\%$.
In a human study, we find that \templm{}'s templates substantially improve upon human-written ones in BERTScore.

\end{abstract}

\section{Introduction}
\begin{figure}[t]
\includegraphics[width=\columnwidth]{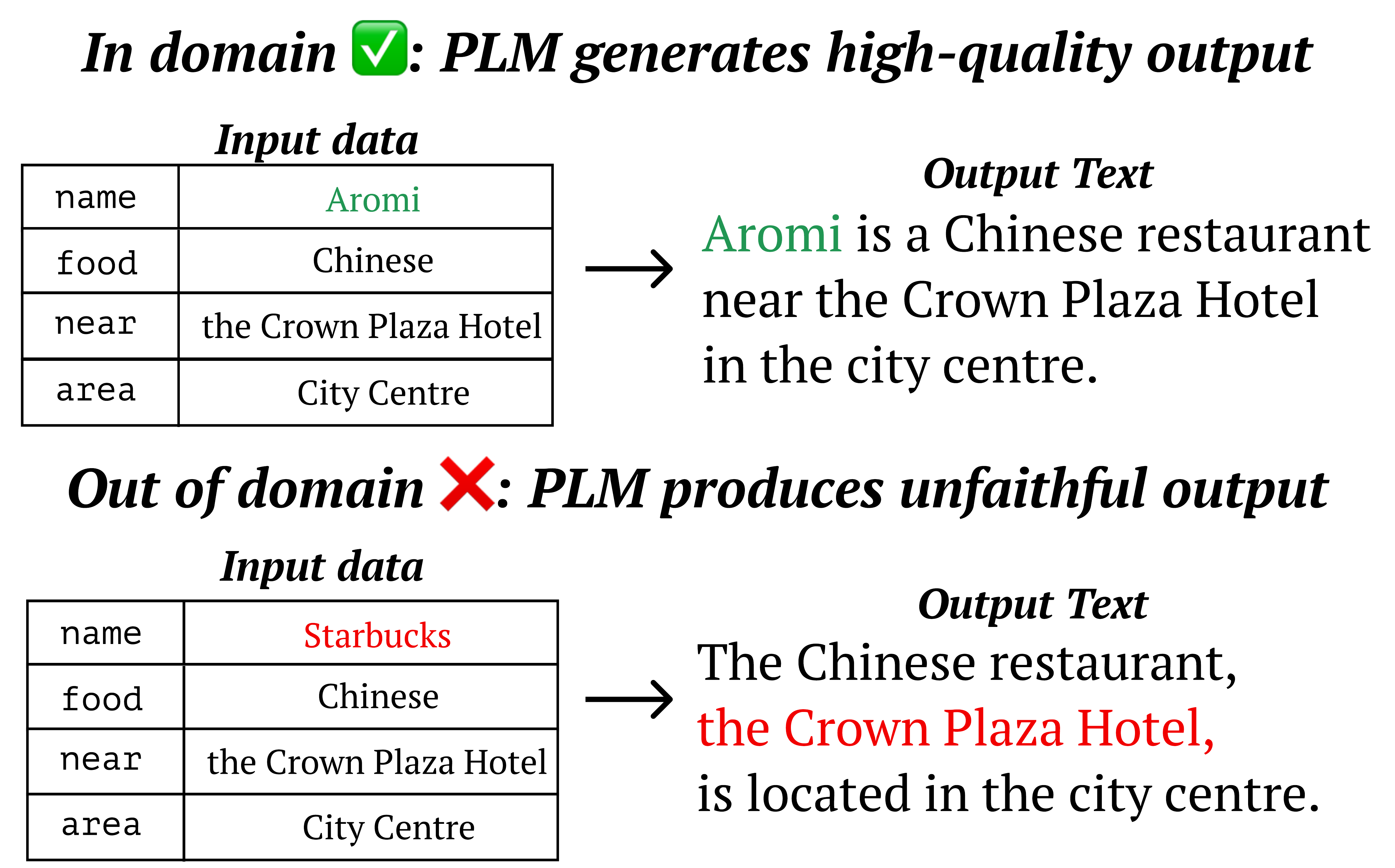}
\caption{
A high-performance \lm{} finetuned on the E2E dataset generates unfaithful outputs when given out-of-domain inputs.
We show later that BART produces such errors $83\%$ of the time while \templm{} never suffers from such failures.
}
\label{fig:fig1-teaser}
\vspace{-10pt}
\end{figure}

\blfootnote{${}^{*}$: Equal Contribution}
Pretrained language models~\citep[\lm{}s;][]{BrMa20lan, Lewis2020BARTDS} can generate fluent text and are data-efficient when being transferred to downstream tasks~\citep{chen-etal-2020-kgpt, Schick2021FewShotTG}.
However, \lm{}s have been known to produce unfaithful outputs~\citep{durmus-etal-2020-feqa, maynez-etal-2020-faithfulness, xiao-wang-2021-hallucination} and inappropriate content~\citep{gehman-etal-2020-realtoxicityprompts} that can lead to disastrous outcomes in real-world deployments~\citep{dungeon-misuse}.
These errors can be worsened when models are queried with out-of-domain (OOD) input.
\Cref{fig:fig1-teaser} shows that querying a finetuned \lm{} with a novel entity (e.g. Starbucks) not in the training data can lead to surprising failures even though the \lm{} achieves high in-domain performance.
This poses a great challenge in deploying \lm{}s in real-world applications.

In stark contrast, classic template-based systems~\citep{reiter_dale_1997, barzilay-lee-2003-learning, angeli-etal-2010-simple} employ templates consisting of words and nonterminal fields, which are robust to novel entities by design.
Moreover, templates are directly readable by humans, and human inspection can provide direct guarantees of faithfulness.
However, templates can be too rigid and produce disfluent text with unexpected inputs.
In this work, we seek to borrow the merits of classic template-based techniques to improve faithfulness and interpretability, while retaining the \lm{}'s flexibility and data efficiency.


We propose \templm{}, a novel framework that \emph{distills} a \lm{} into a template-based system for data-to-text tasks.
At training time, \templm{} extracts templates that maximally recover the induced probability distribution of the \lm{}, similar to model distillation~\citep{Hinton2015DistillingTK}.
At inference time, \templm{} uses the \lm{} to select appropriate data (content selection) and templates (surface realization).\looseness=-1
While distilling a PLM into a template-based generator brings benefits, it also raises new challenges. 
Extracting templates that match a \lm{} is a challenging combinatorial optimization problem with no clear solution. 
Our approach relies on two new ideas.
First, because our goal is to recover the \lm{}'s induced probability distribution, \templm{} initializes its search procedure by \emph{delexicalizing} \lm{}'s generation outputs, \emph{i.e.} abstracting the value in the output with data fields. 
For example, we can delexicalize ``Aromi is a Chinese restaurant'' into ``\texttt{[name]} is a \texttt{[food]} restaurant.''
Second, \templm{} leverages the \lm{}'s generation ability to refine templates, using a novel \textit{consensus beam search} algorithm.
Unlike prior works~\citep{wiseman-etal-2018-learning, wiseman-etal-2021-data}, our approach can leverage any \lm{} to generate templates, allowing us to take advantage of improvements in the data efficiency and fluency of \lm{}s.

We evaluate \templm{} on the E2E~\citep{novikova-etal-2017-e2e} and the SynthBio datasets~\citep{synthbio}.
We observe that \templm{} is the most faithful generation method (with zero faithfulness errors) on E2E and \templm{} fixes the unreliable OOD behavior of \lm{}s, reducing the unfaithful output rate from $83\%$ to $0\%$.
In addition, We show that \templm{} achieves higher metric scores than classic text generation techniques and a previous hybrid neural-template method ($5$ BLEU scores higher than \citet{wiseman-etal-2018-learning} even when trained with $42$ times less data).
We further conduct a human study where we ask annotators to write templates for SynthBio. We observe that \templm{} produces more fluent templates than both the average template writer and an ensemble aggregating all the template writers.
The code for \templm{} is available at \href{https://github.com/Tiiiger/templm}{\tt https://github.com/Tiiiger/templm}.


%

\section{Related Works}

\noindent\textbf{PLMs for language generation.}
\lm{}s~\citep{RaWu19lan, BrMa20lan, Lewis2020BARTDS} are pretrained over large scale text corpus and have significantly improved generation fluency and data efficiency. 
However, much like non-pretrained neural LMs, \lm{}s can produce unreliable outputs, including hallucination~\citep{maynez-etal-2020-faithfulness}, inconsistency~\citep{elazar-etal-2021-measuring}, toxicity~\citep{gehman-etal-2020-realtoxicityprompts}, or privacy violations~\citep{carlini-privacy}. 
\templm{} addresses these shortcomings by distilling a \lm{} into a less expressive but more trustworthy template-based system, while retaining fluency and data efficiency.

\noindent\textbf{Classic template-based methods.}
Classic template methods often delexicalize the training set data, \emph{i.e.,} they abstract the values in examples from the training data with the nonterminal data fields~\citep{Ratnaparkhi2002TrainableAT, oh-rudnicky-2000-stochastic, Rudnicky1999CreatingND, angeli-etal-2010-simple}.
For example, ``The restaurant name is Aromi'' can be delexicalized into ``The restaurant name is \texttt{[name]}.''
However, delexicalization can be challenging for human-written text. 
When describing that the customer rating is ``3 out of 5,'' human writers may paraphrase it into ``3 stars'' or ``average.'' 
Delexicalization has difficulties capturing this paraphrasing problem and leaves lexicalized values in templates, which makes the templates less generalizable.
In contrast, \templm{} first finetunes a \lm{} on the data-to-text task and then exploits the \lm{}'s ability in smoothing the text distribution to tackle the paraphrasing problem.
This technique enables \templm{} to generate more fluent outputs than classic template-based systems.

\noindent\textbf{Hybrid neural generation methods.} There have been a number of works that explore different ways to leverage intermediate representations/operations to guide neural generation, including designing an explicit planning module~\citep{Puduppully2019DatatoTextGW}, editing exemplar training examples~\citep{wiseman-etal-2021-data}, and inducing latent variables~\citep{wiseman-etal-2018-learning, li-rush-2020-posterior, Ye2020Variational}.
Much like classic template-based methods, these systems attempt to learn structured representation from diverse human-written text, which is challenging and often requires heuristics for additional supervision.
We differ from prior methods in two important aspects: 
first, \templm{}'s templates consist of terminal words and nonterminal fields, which make the templates robust and interpretable
second, \templm{} can leverage any \lm{} to generate templates, allowing us to take advantage of improved fluency and data efficiency brought by \lm{}s. \looseness=-1

\begin{figure*}[ht!]
\centering
\includegraphics[width=0.9\textwidth]{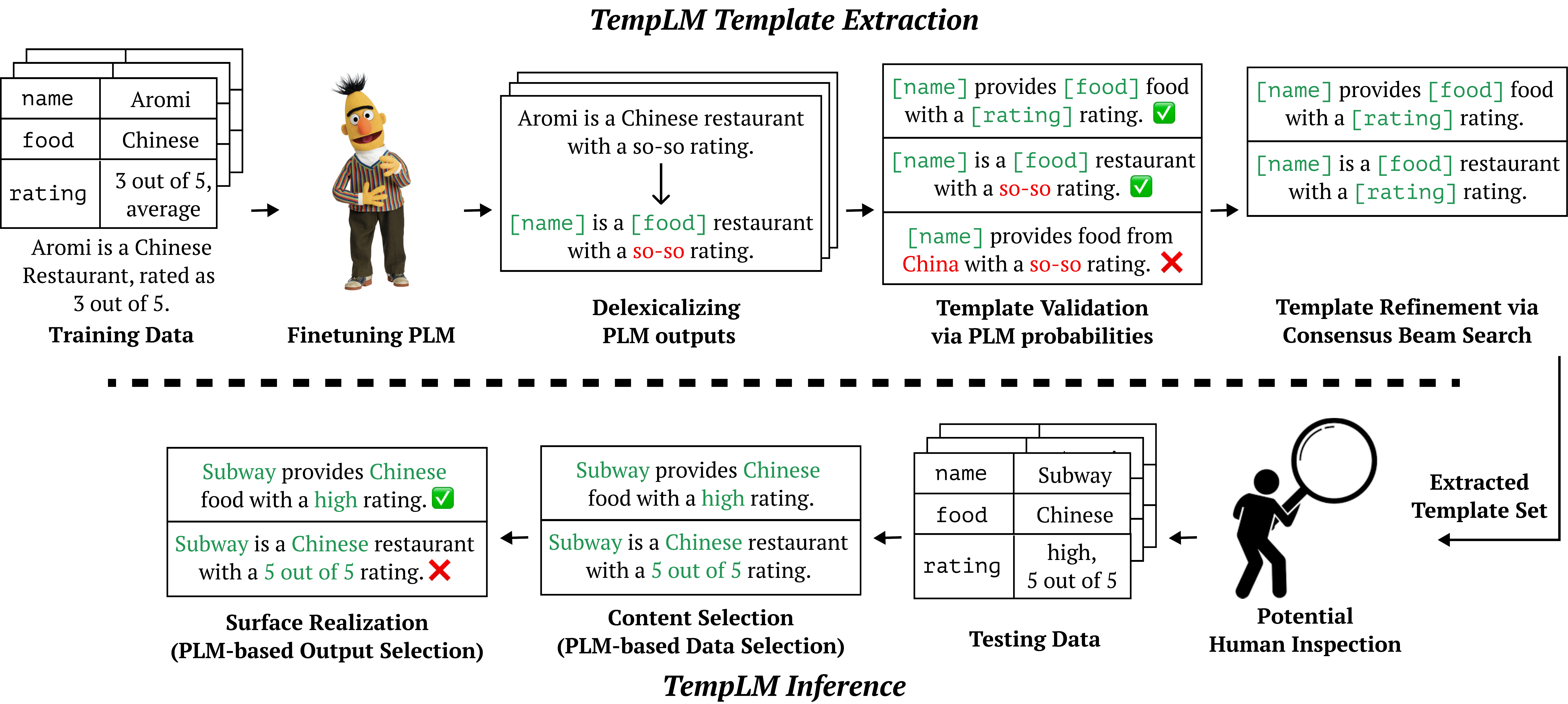}
\caption{Overview of \templm{}. \templm{} performs template extraction and inference by treating the finetuned PLM as the ground truth optimization target. We want to extract generalizable templates that contain nonterminal {\color{applegreen} data fields} and do not contain {\color{red} lexicalized values}.}
\label{fig:system-overview}
\vspace{-10pt}
\end{figure*}

\begin{figure}[t]
    \centering
    \includegraphics[width=0.9\linewidth]{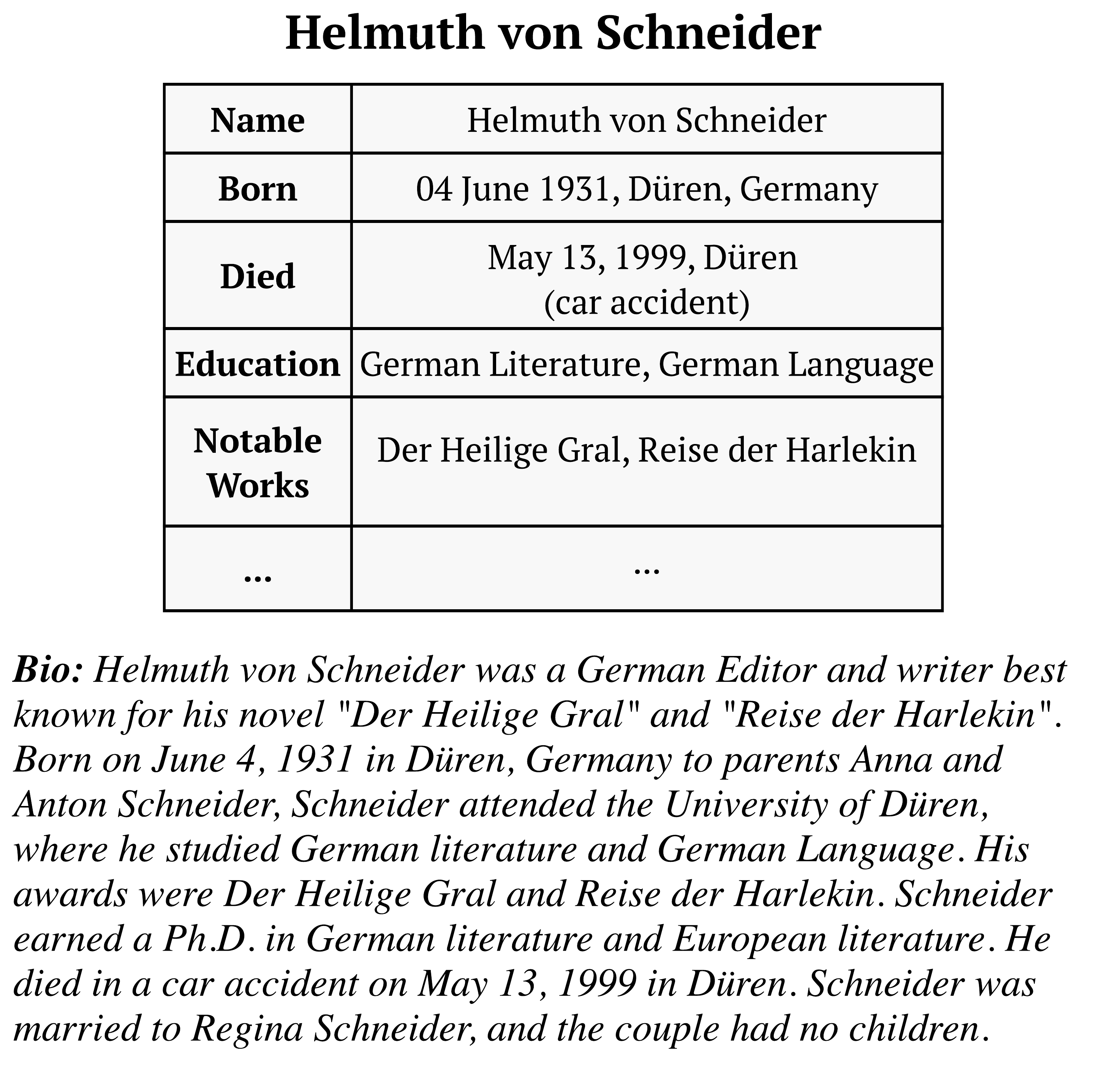}
    \vspace{-10pt}
    \caption{Example of the SynthBio data-to-text task. 
    We are given wikipedia-style data $d$ about a person and are tasked with generating the biography $x$.}
    \label{fig:synthbio-example}
    \vspace{-10pt}
\end{figure}

\section{\templm{}: Template-Based Generators}

\subsection{Problem Statement}
We are interested in data-to-text tasks (\Cref{fig:synthbio-example}), where we are given input data $d$, consisting of \textit{field} and \textit{value} pairs where a field may correspond to multiple values. 
For example, $d$ could be \texttt{\{name: [Aromi, aromi], article: [a, an]\}}, where \texttt{name} is a data field corresponding to multiple values ``Aromi'' and ``aromi''.
Note that we differ from common data-to-text setups in allowing multiple data values and augmenting $d$ with different capitalization and function words to accommodate for template systems.
Our task is to describe $d$ by some text $x \sim p(x|d)$.To this end, we want to learn a model $p_{\theta}(x|d)$ using training examples $(x, d)$.
In the \lm{} approach, $p_{\theta}$ is implemented by finetuning a \lm{} on $(x, d)$, using standard log-loss minimization.

In template-based generation, we want to obtain a template set $T$ consisting of templates $t$ and ensure that for new input data $d$, we can generate a high-quality output $x$.
We define a template $t$ as a sequence of \emph{terminal} tokens and \emph{nonterminal} fields that can be replaced by their values in $d$.
For example, a template ``The restaurant name is \texttt{[name]}'' can be filled in as ``The restaurant name is Aromi''.
We represent the action of filling in a template $t$ with data $d$ as $x = F(t, d)$.

A set of templates $T$ captures the data distribution well if at least one template from $t$ is high-quality for every input $d$. We formalize this
goal by stating that for a given input $d$, we are interested in maximizing $\max_{t\in T} \log p(F(t, d) | d)$. 
Because \templm{} also aims to be inspectable by humans, we want to limit the size of $T$ by a budget $B$, $\lvert T\rvert \leq B$.
Putting these constraints together, we have the following optimization problem:
\begin{equation} \label{eq: objective}
  \argmax_{T, \lvert T\rvert \leq B} \; \E_{d} [ \max_{t \in T} \; \log p(F(t, d) | d)].
\end{equation}

What are the implications of \Cref{eq: objective}?
\Cref{eq: objective} suggests that we would prefer \textbf{generalizable templates} such that a single $t$ can be flexibly filled in so that $\log p(F(t, d)|d)$ is high for many different $d$.
In practice, this means that our objective prefers templates with few or no \emph{lexicalized values}.
Compare the two templates, ``The restaurant name is Aromi'' versus ``The restaurant name is \texttt{[name]}''.
\Cref{eq: objective} would prefer the latter template because the first one does not work well when $d$ describes a different restaurant name.

Although \Cref{eq: objective} nicely captures our intuition of a good template, it presents several optimization challenges. 
\Cref{eq: objective} is a size-constrained combinatorial problem that does not have a clear solution. 
Analyzing the structure of \Cref{eq: objective}, we can decompose it into two separate maximization problems.
First, we have the \textbf{template extraction} problem of identifying the best template set $\argmax_{T, \lvert T\rvert \leq B}$.
Second, given a template set $T$, we have the \textbf{template inference} problem of identifying the best template $\max_{t \in T}$.
We next discuss how to leverage \lm{}s to solve these two problems respectively.

\subsection{Template Extraction}
\label{sec:template-extraction}
The inherent challenge of template extraction is that human-written text in the form of $x \sim p(x|d)$ may not follow a template structure.
This is especially true when humans paraphrase the same data value into different phrasings, but it could also occur as human-written texts have complex syntactic structures that are not covered by templates. 
This linguistic diversity makes delexicalization and more generally learning templates from $x$ extremely challenging.

Our objective in \Cref{eq: objective} resolves this key problem. 
Maximizing $\log p(F(t,d)|d)$ is equivalent asking for a template $t$ to match \emph{any} high probability sequence under $p$, rather than matching \emph{all} high probability sequences, as is typical in delexicalization or latent-variable based template models.
While this approach resolves the paraphrasing problem, it relies upon
 the true data-generating probability $p(F(t,d)|d)$ which we cannot evaluate. 
 Therefore, we propose to approximate $p$ with a PLM $p_{\theta}$. 
This amounts to treating $p_\theta$ as the ground truth optimization target, similar to model distillation~\citep{Hinton2015DistillingTK}.

While targeting $p_{\theta}$ makes the optimization problem easier, \Cref{eq: objective} is still intractable because of its difficult combinatorial structure.
We design a series of approximations to circumvent the optimization difficulty (\Cref{fig:system-overview}).

\paragraph{Clustering.}
Suppose we can obtain the optimal template set $T^{*}=\{t^{*}_1, \ldots, t^{*}_i, \ldots, t^{*}_{B}\}$.
Then we can identify a cluster function $C^{*}$ where $C^{*}(d)=i$ returns the index of the optimal template $t^{*}_{i}$  for example $d$.
With $C^{*}$, we can decompose \Cref{eq: objective} into $B$ sub problems that are easier to solve,
\begin{equation} \label{eq: decompose-objective}
  \argmax_{t_i} \; \mathop{\E}_{d \text{ s.t. } C^{*}(d)=i} [ \log p_{\theta}(F(t_i, d) | d)].
\end{equation}
While obtaining $C^{*}$ is impossible, we can design approximate clusters $C$ based on the presence of different fields, as is standard in other data-to-text methods \citep{wiseman-etal-2021-data}.

\paragraph{Delexicalizing \lm{} outputs.}
Equipped with approximate clusters $C$, how can we find templates that work for all examples in the same cluster?
Because we are optimizing for $p_{\theta}$, one natural starting point is to delexicalize the model beam search output $x_{\theta}$.
We denote \delext{} as the template we obtain from delexicalizing the \lm{} output $x_{\theta}$ of the input $d$ and \delexT{} the template set.

Delexicalizing $x_{\theta}$ also allows us to be more data efficient and potentially more robust to out-of-domain scenarios.
This is because obtaining \delexT{} only requires unlabeled inputs $d$ as opposed to requiring full supervision $(x, d)$, which allows us to exploit data beyond the training set.
In practice, we perform data recombination~\citep{jia-liang-2016-data} to not only increase the quantity of $d$ but also explore more field and value compositions.

\paragraph{Template validation via \lm{} probabilities.}
While \delexT{} provides a good initial template set, some of these templates may contain a substantial number of lexicalized data values.
To filter these less generalizable templates and fulfil the template budget constraint $B$, we want to filter the template set $\lvert T^{\texttt{delex}}_{\theta}(d) \rvert$.
We leverage the \lm{}'s soft probabilities measure to evaluate the template \emph{generalizability}, defined as a template's average log probability over the entire cluster.
For a template generated by delexicalizing $d$, this objective can be written as 
\begin{equation} \label{eq: pruning-objective}
  \mathop{\Sigma}_{d' \text{ s.t. } C(d')=C(d)} [ \log p_{\theta}(F(t^{\texttt{delex}}_{\theta}(d), d') | d')].
\end{equation}
where $d'$ are examples sampled from the same data cluster, $C(d')=C(d)$.
\Cref{eq: pruning-objective} assigns a scalar value to each \delext{} and by ranking this value, we filter out any brittle templates.
In practice, we retain the top-$K$ best templates in each cluster to form the template set.

\paragraph{Template refinement via Consensus Beam Search.} 

\begin{figure}[t!]
\renewcommand{\figurename}{Algorithm 1}
\renewcommand{\thefigure}{}
\algrenewcommand{\algorithmiccomment}[1]{\hskip1em\# \footnotesize#1 \small}
\begin{algorithm}[H]
\small
$k$: beam size, \ $M$: maximum length\\
$\vocab$: terminal tokens, $\vocab_{T}$: nonterminal fields \\
$N$: number of inputs \\
$t'$: partial template where ungeneralizable spans are removed \\
$x'_{i}$: $F(t', d_{i})$ , $d_{i}$: $i$th input data \\
$d_{i}.\mathrm{get}(\cdot)$: return the best value token for a field token
\begin{algorithmic}[1]
\State $B_0 \gets \{ \langle 0, \bos \rangle \}$
\For{ $t \in \{ 1, \dots, M\!-\! 1 \}$ }
    \State $H \gets \varnothing$
    \For{$ \langle s, \rvy \rangle \in B_{t-1}$} \algorithmiccomment{Expansion.}
    \For{$y \in \vocab \circ \vocab_{T}$}
    \State $ S \gets \varnothing$ \label{line: agg-start}
    \For{$i \in \{ 1, \dots, N\!-\! 1 \}$}
        \If{$y \in \vocab$}
            \State $S.\mathrm{add}( \log p_{\theta}(\rvy \circ y | x'_i, d_i))$
        \Else
            \algorithmiccomment{Field token substitution.}
            \State $S.\mathrm{add}( \log p_{\theta}(\rvy \circ d_i.\mathrm{get}(y) | x'_i, d_i))$ \label{line: field-sub}
        \EndIf
    \EndFor
    \State $s \gets S.\mathrm{avg}()$ \algorithmiccomment{Aggregation.} \label{line: agg-end}
    \State $H.\mathrm{add}( \langle s, \rvy \circ y \rangle)$
    \EndFor
    \EndFor
 \State $B_t \gets H.\mathrm{topk}(k)$
\EndFor
\State \Return $B_t.\mathrm{max}()$

\end{algorithmic}
\caption{\small \textbf{Consensus Beam Search}}
\label{alg:beam-search}
\end{algorithm}
\caption{We search for a common constituent $\rvy$ that can be infilled to all partial descriptions $x'_i$.
In contrast to conventional beam search, we aggregate the log probability scores across different inputs at each step in \Cref{line: agg-start} to \Cref{line: agg-end}.
To generate nonterminal fields (\emph{e.g.} \texttt{[name]}), we account for how they will be filled in with different input $d'_i$ in \Cref{line: field-sub}.
}
\end{figure}

If a template contains only a few lexicalized values, we could identify these spans using a token-level version of \Cref{eq: pruning-objective} and then replace ungeneralizable spans by executing a search algorithm with \Cref{eq: pruning-objective} as the objective.

To identify the non-generalizable spans, we begin by evaluating the token-level equivalent to \Cref{eq: pruning-objective} (see \Cref{sec:app:tok-general} for details).
We then aggregate these token-level scores into a constituent-level score using a constituency parser, and mark any constituent whose generalizability is worse than a threshold as ungeneralizable.

To salvage these ungeneralizable spans, we leverage a \lm{} to optimize for \Cref{eq: pruning-objective} directly.
We remove the ungeneralizable spans to form partial template $x'$ and learn an infilling model $p_{\theta}^{\text{infill}}(x|x', d)$ to replace the ungenearlizable spans.
We implement $p_{\theta}^{\text{infill}}$ by finetuning a different \lm{} and present the details in \Cref{sec:app:in-domain}.

There are two challenges we face in optimizing \Cref{eq: pruning-objective}. 
First, the infilling model $p_{\theta}^{\text{infill}}$ is learned to generate text, not templates. 
Second, \Cref{eq: pruning-objective} is an unusual objective in text generation that is a mixture-of-experts of many language models where each model conditions on some input $d'$.
We propose two modifications to the standard beam search algorithm to address these challenges (\Cref{alg:beam-search}).
First, we empower the infilling model $p_{\theta}^{\text{infill}}$ with the ability to generate nonterminal data fields and define their scores based on how they will be filled in (\Cref{line: field-sub}).
Second, we search for a common output that is the ``consensus'' of many inputs $d'$ by aggregating the log probability scores across inputs at each decoding step (\Cref{line: agg-start} to \Cref{line: agg-end}).
Empirically we find that template refinement can correct for errors in the earlier steps and remove lexicalized values in the template or incorrect fields in the template. We present a qualitative study of template refinement in \Cref{sec:qualitative}.
\label{subsec:infer}

\paragraph{Human Inspection and Validation.}
Once templates are refined, we save them as an internal part of \templm{} and use them later for template inference.
To obtain an even stronger faithfulness guarantee, we can have human inspectors validate each template.
\templm{} offers two advantages for such human-in-the-loop inspection.
First, templates in \templm{} are directly readable by humans.
Second, \templm{} by design has very limited freedom during inference: 
an output can only be generated from filling in a template with input data.
As long as none of the templates contains hallucination or inconsistency, \templm{} will be guaranteed to return a faithful output.
The combination of interpretability and restricted output space
 enables a natural interface for human-in-the-loop cooperation, where a human inspector can sanitize all the templates before deploying \templm{} into real-world applications.

\subsection{\templm{} Template Inference}
Given the template set $T$ that we extracted, we now need to solve the problem of identifying the best template $\max_{t \in T}$ for a new input $d$.
In \templm{}, we leverage \lm{}s as a core primitive in both the content selection and surface realization steps.

\textbf{Content Selection} requires us to substitute a nonterminal field with the most appropriate value among the multiple values that a field corresponds to.
We perform this step using a left-to-right auto-regressive \lm{}.
At each decoding step, we directly copy from $t$ when encountering a terminal word; otherwise, we select the most probable data value to replace a field.
\lm{}s are typically trained with byte-pair encoding~\citep{sennrich-etal-2016-neural}, which might break up data values into multiple tokens.
Performing an exact search involves computing the probability of each multi-token value by additional roll-outs, which slows down inference.
We circumvent this problem by performing a greedy search on the first token, which leads to faster or on-par inference time with standard \lm{} inference.

\textbf{Surface Realization} requires us to select the most appropriate output after templates are filled in.
We perform this step by computing $F(t, d)$ for all templates in the same cluster $C(d)$ and returning the one with highest $p_{\theta}(F(t,d)|d)$.

\section{Experiments}
We evaluate \templm{}'s ability to generate faithful and fluent text in three settings:
an in-domain evaluation on standard data-to-text benchmarks, an out-of-domain evaluation that stress tests the ability to novel input, and a human study comparing \templm{}'s template extraction ability to that of human template writers.

\subsection{Experiment Setup}
\begin{table}[]
    \centering
    \begin{tabular}{c|ccc}
    \toprule
         &  \# Train & Average Length & \# Fields\\
    \midrule
         E2E &  $1090$ & $19.8$ & $8$\\
         SynthBio & $2896$ & $93.1$ & $78$\\
    \bottomrule
    \end{tabular}
    \caption{Statistics of SynthBio and the downsampled E2E dataset.}
    \label{tab:data_stats}
    \vspace{-10pt}
\end{table}
\begin{table*}
\setlength\tabcolsep{3pt} 
\begin{tabular}{lr}
\begin{minipage}{.5\linewidth}
\small
\centering
\begin{tabular}{l|cccc}
\toprule
 & $E_{\text{precision}}\downarrow$  & \bleu $\uparrow$& \rougel $\uparrow$\\
\midrule
BART & $6.0\pm 2.9$ & $\mathbf{66.2}\pm 0.5$ & $\mathbf{68.4}\pm 0.7$\\
\midrule
TempLM & $\mathbf{0.0}\pm 0.0$ & $61.5\pm 1.0$ & $64.5\pm 0.8$\\
\midrule
$\text{NTemp}^{\dagger}$ & $7$ & $55.17$ & $65.70$\\
TempClassic & $46.7\pm 25.4$ & $52.1\pm 2.0$ & $62.2\pm 2.3$\\
SUB & $110.7\pm 36.2$ & $45.3\pm 1.9$ & $55.6\pm 2.4$\\
\bottomrule
\end{tabular}
\end{minipage}&
\begin{minipage}{.5\linewidth}
\centering
\small
\begin{tabular}{l|cccc}
\toprule
 & \bleu $\uparrow$& \bertf $\uparrow$\\
\midrule
BART & $\mathbf{40.8}\pm 0.2$ & $\mathbf{55.2}\pm 0.1$ \\
\midrule
TempLM & $40.3\pm 0.3$ & $54.3\pm 0.1$  \\
\midrule
TempClassic & $36.6\pm 0.2$ & $48.8\pm 0.1$ \\
SUB & $14.1\pm 0.1$ & $18.9\pm 0.1$ \\
\bottomrule
\end{tabular}
\end{minipage}
\end{tabular}
\caption{
  Automatic metrics averaged over three random seeds on the E2E and SynthBio test sets.
  We bold the best numbers in each column and show standard errors with error bars.
First, we observe that \templm{} produces zero unfaithful outputs on E2E.
Second, \templm{} achieves better or on-par performance on reference-based evaluation than other template systems. 
${}^{\dagger}$:We compare to a model trained on the \textit{full} E2E training set, which was released by \citet{wiseman-etal-2018-learning}. 
We were unable to train NTemp models on the subsampled E2E dataset to convergence.
}
\label{tab:auto-eval}
\end{table*}

\begin{table*}[]
\begin{center}
\small
\begin{tabular}{lcccccc}
\toprule
& \multicolumn{4}{c}{E2E} & \multicolumn{2}{c}{SynthBio}\\
\cmidrule(r){2-5}
\cmidrule(l){6-7}
& $E_{\text{precision}} \downarrow$ & \%. Lex. Temp $\downarrow$ & \bleu $\uparrow$ & \#. Temp $\downarrow$ & \bleu $\uparrow$ & \#. Temp $\downarrow$\\
\midrule
TempLM & $\mathbf{0.0}\pm 0.0$ & $\mathbf{5.2}\pm 1.2$ & $61.5\pm 1.0$ & $\mathbf{471.7}\pm 62.9$ & $\mathbf{40.3}\pm 0.3$ & $\mathbf{80}$\\
- Refinement & $\mathbf{0.0}\pm 0.0$ & $12.1\pm 1.3$ & $61.4\pm 0.9$ & $534.3\pm 8.5$ & $35.2\pm 0.9$ & $\mathbf{80}$ \\
- Validation & $2.7\pm 2.2$ & $21.4\pm 2.6$ & $\mathbf{64.0}\pm 1.0$ & $2047.3\pm 43.7$ & $36.4\pm 0.1$ & $1511$ \\
TempClassic & $46.7\pm 25.4$ & $37.4\pm 0.5$ & $52.1\pm 2.0$ & $978.3\pm 1.2$ & $36.6\pm 0.2$ & $1511$ \\
\bottomrule
\end{tabular}
\end{center}
\vspace{-5pt}
\caption{
Ablation results averaged over three random seeds on different template-based systems. 
We bold the best numbers in each column and show standard errors with error bars.
\templm{} extracts most generalizable templates and achieves good performance with a small number of templates.}
\label{tab:ablation}
\vspace{-10pt}
\end{table*}

\paragraph{Datasets.}
We consider two data-to-text datasets: E2E~\citep{novikova-etal-2017-e2e} and SynthBio~\citep{synthbio}.
The E2E dataset contains data entries about restaurants and asks for text descriptions of restaurant data.
Originally, the E2E dataset contained $42K$ training samples with eight distinct fields and $109$ field combinations.
To better evaluate data efficiency and faithfulness, we downsample the training set to ten samples per field combination.
Results on the full E2E dataset are similar and are shown in \Cref{sec:app:in-domain}.
We evaluate on the official validation and test sets.

SynthBio asks systems to write biographies based on Wikipedia-style data tables and was originally proposed as an evaluation set for WikiBio~\citep{wikibio}.
Because WikiBio is a noisy dataset created by automatic retrieval and contains pervasive hallucinations, we avoid using it.
Instead, we split SynthBio into training, validation, and test sets, and evaluate on the test set. 
We summarize the dataset statistics in \Cref{tab:data_stats}.

\paragraph{Evaluation Metrics.}
We evaluate fluency of the generated outputs by reference-based evaluation. For E2E, we use the official toolkit and evaluate in terms of \bleu~\citep{bleu}, NIST~\citep{nist}, \rougel~\citep{rouge}, CIDEr~\citep{cider}, and METEOR~\citep{meteor}.
For SynthBio, we evaluate by \bleu, \rougel, and BERTScore~\citep{bert-score}.

On the E2E dataset, we also evaluate the faithfulness of a system output. 
We define an output description to be faithful if it does not contradict the input data or hallucinate information not present in the input. 
To automatically evaluate this, we manually inspected system output descriptions in the validation set and collected common paraphrases of each possible data value.
For example, a customer rating of ``3 out of 5'', may appear as ``3 stars'', ``average'', etc.
This allows us to develop a matching-based metric: we count precision error \underline{$E_{\text{precision}}$} when a piece of system output contains any paraphrase that matches with a value not in the input (hallucination) or a value different from the one provide in the input (inconsistency). 

Note that $E_{\text{precision}}$ is a conservative metric.
When we encounter novel phrasings that does not match any entry in our phrasing collection, we do not count them toward $E_{\text{precision}}$ and only measure cases where we are certain that an output contains hallucination or inconsistency.
We present more implementation details in \Cref{sec:app:hypers}.
For template-based methods, we reuse the same routine to measure the percentage of templates that contain lexicalized values ({\underline{\%. Lex. Temp}}), which measures the generalizability of the generated templates.
We calculate an analogous recall-oriented metric $E_{\text{recall}}$ but because E2E does not require systems to verbalize every value in $d$, we do not focus on this metric and include the results in \Cref{sec:app:in-domain}.

\paragraph{Implementing \templm{}.}
We implement $p_{\theta}(x|d)$ and the infilling model $p_{\theta}(x|x', d)$ by finetuning $\text{BART}_{\text{BASE}}$~\citep{Lewis2020BARTDS} models.
On E2E, we assign data that has the same combination of fields into the same cluster, which results in $109$ clusters.
We use data recombination~\citep{jia-liang-2016-data} to combinatorially create $50$ samples for each cluster and thereby increase the training data size by five times for template extraction.
We define the target number of templates per cluster for \templm{} to be five, which results in around $500$ templates after deduplication.
On SynthBio, we cluster data by the ``occupation'' field, which results in eight clusters, and we target \templm{} for ten templates per cluster.
We do not perform any data augmentation for SynthBio.
More training details are described in \Cref{sec:app:hypers}.

\paragraph{Baselines.}

\looseness-1 We compare to three classes of baselines over three seeded runs. To compare to existing \lm{}s, we compare to a finetuned \underline{$\text{BART}_{\text{BASE}}$} model.

To compare to classic template systems that delexicalize training samples, we compare to \underline{TempClassic}, which delexicalizes the training data but uses our \lm{}s based inference procedure. 
We also compare to the \underline{SUB} baseline~\citep{wiseman-etal-2018-learning}, which replaces the \lm{}s based inference in TempClassic with a rule-based procedure.

To compare to recent hybrid neural-template methods, we compare to the \underline{NTemp} method  \citep{wiseman-etal-2018-learning}.
As we were unable to obtain good performance by NTemp on the downsampled training set, we evaluate the model trained on the full E2E training set, which was released by \citet{wiseman-etal-2018-learning}.

Finally, we performed ablation studies by removing the template refinement (\underline{- Refinement}) and template validation (\underline{- Validation}) components from \templm{}.

\subsection{In-domain Experiment}
\label{sec:in-domain}

\Cref{tab:auto-eval} shows that on the E2E and SynthBio test sets, \templm{} is more faithful than BART while achieving higher metric scores than other template-based methods.
We present other metric scores and validation set results in \Cref{sec:app:in-domain}.

\paragraph{\templm{} is faithful.}
\templm{} is the only method that achieves \emph{zero} $E_{\text{precision}}$ across validation and test sets.
This improvement over BART suggests \templm{}'s usefulness in practice.
For real-world deployments, we can further leverage human inspection to sanitize \templm{}'s template set, which allows us to remove any lexicalized values in the templates and obtain strict guarantees for \templm{}'s faithfulness.
In contrast, TempClassic produces almost eight times more precision errors than BART ($46$ vs. $6$), which shows the difficulty of inducing templates over human-written text.

\paragraph{\templm{} is fluent.}
We observe that \templm{} achieves higher metric scores than classic template baselines and NTemp~\citep{wiseman-etal-2018-learning}, and on SynthBio, \templm{} even performs similarly to BART despite using the less expressive template representation.
This demonstrates that \templm{} achieves better fluency than previous template methods and validates our ideas of leveraging \lm{}s for template extraction.
In particular, we note that \templm{} achieves a significant $5$ BLEU score improvement over NTemp, which is trained with much more data ($1090$ vs. $42K$ training samples).
This comparison shows that \templm{} is able to retain the impressive data efficiency of \lm{}s.

\paragraph{\templm{} enables trade-offs between fluency, robustness, and interpretability.}
We designed \templm{} to have a small number of templates to make \templm{} more conducive to human inspection.
\templm{} successfully achieves this, using less than $500$ templates for E2E and only $80$ templates for SynthBio.
Comparing \templm{} without Refinement and \templm{} without Validation, we find that template validation reduces the number of templates and substantially increases reliability (halving the percentage of templates containing lexicalized values), but may incur a minor performance drop in fluency.

We find that the template structure is simpler on E2E, and refinement does not add substantial benefit.
However, refinement results in dramatic gains in SynthBio, where it is critical to reversing the performance drop from template validation and results in a $4$ BLEU score gain.
Upon inspection, we found that template refinement can accurately remove ungeneralizable spans in the longer and more complicated templates required for SynthBio. 

Overall, we find that \templm{} ensures faithfulness, retains the \lm{}'s fluency and data efficiency, and balances between performance and interpretability.
In the following sections, we go beyond automatic in-domain evaluation.
We first stress test systems with out-of-domain input, perform a human study to showcase the difficulty of template extraction, and finally analyze a qualitative example to illustrate template refinement.

\subsection{Out-of-domain Experiment}
\begin{table}
\begin{center}
\begin{tabular}{c|c}
\toprule
& Unfaithful Output Rate (\%) \\
\midrule
BART & $83.3$\\
TempLM & $0$\\
\bottomrule
\end{tabular}
\end{center}
\caption{Human annotated unfaithful output rates in out-of-domain (OOD) evaluation.
We observe BART outputs exhibit pervasive unfaithful errors whereas TempLM continues to remain faithful.
}
\vspace{-10pt}
\label{tab:e2e_ood}
\end{table}

Models deployed in real-world applications often encounter test distributions different from the training distribution.
For example, a data-to-text model for restaurant descriptions should work well for new restaurant names not in the training set.
To test for out-of-domain (OOD) generalization, we simulate such a setting on E2E and evaluate BART and \templm{} on OOD input.

We create our OOD evaluation by taking fields in E2E \texttt{(area, eatType, food, name, near)} and filling in common entities scraped from the internet to create $54$ novel examples.
For instance, we create examples like \texttt{\{area: Central Park, eatType: restaurant, food: German, name: McDonald's, near: Subway\}}.
We inspect the system outputs manually to check the correctness and present the results of this study in \Cref{tab:e2e_ood}.
We observe that BART produces unfaithful output frequently, often confusing entities from different types.
In the previous example, BART mistakenly outputs ``Central park is a restaurant ...'', confusing \texttt{area} with \texttt{name}. 
In other cases, BART would ignore the input value and hallucinate entities from the training set, such as ``city centre''.
In contrast, \templm{} is robust to novel inputs and does not produce \emph{any} unfaithful outputs.
We provide the list of novel entities used in creating OOD input and more qualitative examples in \Cref{sec:app:ood}.

\subsection{Human Study}
\label{sec:human}
\begin{table}
\begin{center}
\begin{tabular}{c|l|c}
\toprule
&  & \bertf \\
\midrule
\multirow{5}{*}{\Centerstack{Writer\\ Cluster}} & Human & $51.3\pm 2.3$\\
 & \Centerstack[l]{Human \\Ensemble} & $54.0$\\
 & BART & $58.5\pm 0.2$\\
 & TempLM & $58.8\pm 1.0$\\
\midrule
\multirow{5}{*}{\Centerstack{Spy\\ Cluster}} & Human & $42.2\pm 4.4$\\
 & \Centerstack[l]{Human \\Ensemble} & $48.5$\\
 & BART & $55.3\pm 0.1$\\
 & TempLM & $50.8\pm 0.9$\\
\bottomrule
\end{tabular}
\end{center}
\caption{Human study results on two clusters of the SynthBio test set. Human-written templates result in low metric scores even in the ensemble setting, showcasing the difficulty of identifying distributional characteristics for human and the efficacy of TempLM.}
\label{tab:synthbio-human}
\vspace{-10pt}
\end{table}

\setcounter{figure}{3}

\begin{figure*}[ht]
\centering
\includegraphics[width=0.8\textwidth]{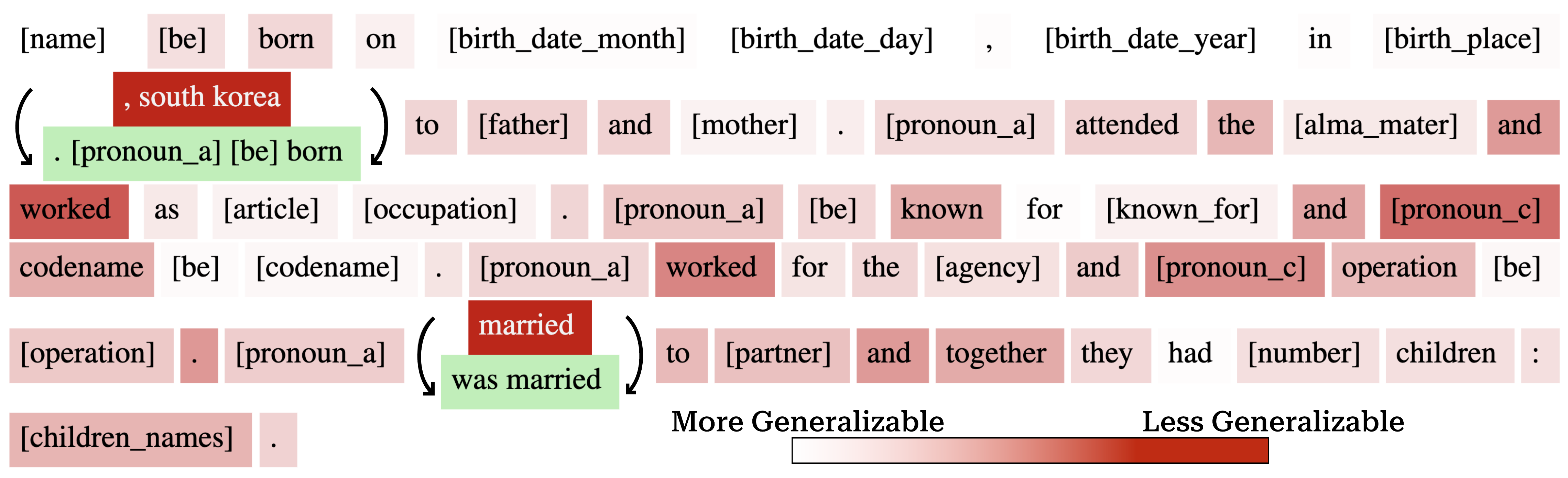}
\caption{A qualitative example of \templm{} refinement process. We color a terminal word or a nonterminal field as more {\color{red}red} if it is less generalizable, measured by the token-level generalizability (\Cref{sec:app:tok-general}). We mark the refinements \templm{} by arrows, coloring the refinement outcome in {\color{applegreen} green}.}
\label{fig:refine-qualitative}
\end{figure*}

To demonstrate the difficulty of generating templates without \templm{}, we conduct a human study on two clusters of the SynthBio dataset.  
We recruited ten volunteers from the CS department to be our template writers and assigned five writers to work on each cluster.
Each template writer was given thirty minutes to write templates, and each template writer wrote eleven templates on average. We presented them the same data that \templm{} operated on: template writers are presented with roughly $200$ training examples, including the input data $d$ and annotated output $x$.
We include our human study instruction and interface in \Cref{sec:app:human}.

To evaluate human performance, we used the human-written templates in our LM-based inference pipeline and measured automatic metric scores. 
\Cref{tab:synthbio-human} shows the BERTScore F1 for both the average template writer as well as an ensemble of five template writers. We report other metric scores in \Cref{sec:app:human}.
We observe that the templates extracted by \templm{} lead to better performance than the human-written ones, indicating the intrinsic difficulty of template writing.
Based on our observation during the template writing process, we found that a common strategy employed by our template writers is to first go through a subset of the training examples and then find canonical examples to delexicalize.
However, we identified a few shortcomings to this procedure.
First, our writers typically only read a few examples (approximately $5$ to $20$) before they exhaust their cognitive load.
As a result, some of our writers fail to write templates that capture the less common examples.
Second, our volunteers may fail to pick the more canonical examples and choose delexicalize examples that are not the most generalizable.
Although well-trained template writers with domain knowledge might have written better templates, the difficulty in identifying such distributional characteristics remains true for any sizable data.

We were also delighted to see that one of our volunteers creatively employed a compositional approach to template writing,
where they wrote templates for each sentence in the biography and efficiently obtained templates by reordering and recomposing different sentence-level templates.
This approach performed well and we hope to include compositionality as an inductive bias for \templm{} in our future work.


\subsection{Qualitative Analysis of \templm{}}
\label{sec:qualitative}

To better explain the inner workings of \templm{}, we visualize one example of refinement in \Cref{fig:refine-qualitative}.
We color each word according to its generalizability, measured by a token-level generalizability (see \Cref{sec:app:tok-general}).
From \Cref{fig:refine-qualitative}, we first observe that our generalizability measure is reliable, successfully distinguishing the lexicalized value ``south korea'' and disfluent span ``married'' from the rest of the template.
Second, we observe that the refinement step correctly fixes both errors by replacing ``south korea'' with more generalizable, nonterminal fields and inserting ``was'' to fix the grammatical error.
\Cref{fig:refine-qualitative} demonstrates the effectiveness of template refinement and helps explain why refinement leads to a substantial performance gain on SynthBio in \Cref{tab:ablation}.

From \Cref{fig:refine-qualitative}, we also observe that the words after ``and'' often appear less generalizable. 
This is because there are many alternative ``branches'' that could continue the prefix in these positions and each alternative option will receive a lower probability under a left-to-right \lm{} $p_{\theta}(x|d)$.
We find that the infilling \lm{} $p_{\theta}(x|x', d)$ is robust to these false positives and typically will leave these spans unchanged.
This illustrates the benefits of combining a left-to-right and an infilling \lm{}s in template refinement.
\section{Conclusion and Future Work}
We propose \templm{}, a novel framework for distilling \lm{}s into template-based systems.
\templm{} is designed to achieve better robustness and interpretability while inheriting the fluency and data efficiency of \lm{}s.
Our evaluations show that \templm{} can completely eliminate the unfaithful outputs produced by a finetuned BART model for out-of-domain inputs.
On in-domain evaluation, \templm{} is able to produce more fluent outputs compared to classic template systems, prior neural-hybrid template methods, and even human template writers. 
In the future, we look forward to extending the \templm{} framework to learn compositional templates and grammars, as well as improving its coverage to diverse outputs, potentially via paraphrases of its input data.

\section*{Acknowledgement}
We thank Alex Wang, Nelson Liu, Kaitlyn Zhou, Xuechen Li, Rishi Bommasani, Niladri Chatterji, Shibani Santurkar, Rohan Taori, and Mirac Suzgun for participating in our human study.
Tianyi Zhang was partially supported by the center for research on foundation models (CRFM).
Lisa Li was supported by the Stanford Graduate Fellowship.
Mina Lee was supported by a PECASE award.
\bibliography{custom}

\appendix
\onecolumn

\section{Additional Details on Template Refinement}
\subsection{Token-level Generalizability Measure}
\label{sec:app:tok-general}
Our goal is to  set of generalizable templates given a budget $B$ such that a single $t$ can be flexibly filled in so that $\log p_{\theta}(F(t, d)|d)$ is high for many different input data $d$.
\Cref{eq: pruning-objective} does this exactly: we fill in a single template $t$ with many other input data $d$ from the same cluster and measure the sum of their log probabilities.
We want generalize \Cref{eq: pruning-objective} to a token-level generalizability measure, which tells us which tokens within a template $t$ will receive high probability after the template is filled in with new data.
Our idea is to align tokens in the template with tokens in the output and aggregate the corresponding token probabilities across many different outputs.

Let us use $j$ as the token index and denote $x_{j}$ as the $j$th token in an output text $x$ and $t_{j}$ as the $j$th token in a template $t$.
We use $x_{:j}$ to represent the prefix up to the $j$th token in $x$ and analogously defined $t_{:j}$.
We leverage an alignment function $A(t, d, j)$, where $F(t, d)_{A(t, d, j)}$ givens the token that correspond to $t_{j}$ after $t$ is filled in.
The alignment $A$ handles the discrepancy in length that is caused by the template fill-in process because the fill-in function $F$ substitutes nonterminal fields with various length data given in $d$.
With the help of $A$, we can define the token-level generalizability for a token $t_j$ as,
\begin{equation} \label{eq: token-general}
  \mathop{\Sigma}_{d' \text{ s.t. } C(d')=C(d)} [ \log p_{\theta}(F(t^{\texttt{delex}}_{\theta}(d)_{A(t, d, j)}, d') | F(t^{\texttt{delex}}, d')_{\theta}(d)_{:A(t, d, j)}].
\end{equation}

\Cref{eq: token-general} provides a token-level measure, which we can easily turn into a span-level measure by calculating the joint token-level probability.
We use this idea to calculate the generalizability of nonterminal fields that correspond to values of multiple tokens.
\Cref{eq: token-general} gives us an useful tool for telling which tokens are ungeneralizable and we can then leverage the generation ability to replace these tokens by directly optimizing \Cref{eq: token-general}.

Now that we formalize token-level gerneralizability with \Cref{eq: token-general}, our plan is to iteratively remove ungeneralizable spans and use an infilling model to generate new template spans.
We can decompose this procedure into two subproblems: removing ungeneralizable spans and generating new template spans.
We discuss them in the next two sections, respectively.

\subsection{Removing Ungeneralizable Spans}
The key problem we want to solve in span removal is to group multiple ungeneralizable tokens together and remove them at the same time.
This is because if we remove ungenralizable token one at a time, we would still condition on other ungenralizable tokens, which deteriorates performance in practice.
We leverage constituency parsing~\citep{kitaev-klein-2018-constituency} to solve this problem.
For each constituent in the parse tree, we calculate \Cref{eq: token-general} for each token in the constituent and compute the average.
We set a threshold and remove all constituents whose generalizability measure is worse than this threshold.

\subsection{Generating Template with Consensus Beam Search}
We refer to \Cref{sec:template-extraction} for the description of our template generation process.
In \Cref{alg:beam-search}, we rely on the subroutine $d_{i}.\mathrm{get}(\cdot)$, which gives us the best data value among the multiple options in $d$ for a nonterminal field.
Implementing this subroutine exactly requires us to evaluate all data values at each decoding step, which is computationally expensive.
In practice, we perform a greedy selection based on the first token in each data value.
\section{Additional Details on Experiments}
\subsection{Model Training Details}
\label{sec:app:hypers}
\paragraph{Left-to-right Autoregressive LM.}
We finetune a $\text{BART}_{\text{BASE}}$ model to implement $p_{\theta}(x|d)$.
On the downsampled E2E dataset, we train for $10$ epochs for a batch size of $16$ and a learning rate of $3\times 10^{-5}$. We train with half precision using the huggingface implementation.
On SynthBio, we train for $5$ epochs for a batch size of $8$ and a learning rate of $3\times 10^{-5}$. We train with half precision using the huggingface implementation.
\paragraph{Infilling LM.}
We train our infilling models by masking a random $0$ to $10$ words span and predicting the masked out span.
We finetune a $\text{BART}_{\text{BASE}}$ model to implement $p_{\theta}(x|x', d)$.
On the downsampled E2E dataset, we train for $50$ epochs for a batch size of $16$ and a learning rate of $3\times 10^{-5}$. We train with half precision using the huggingface implementation.
On SynthBio, we train for $20$ epochs for a batch size of $16$ and a learning rate of $3\times 10^{-5}$. We train with half precision using the huggingface implementation.
\paragraph{\templm{}.}
On E2E, we cluster based on field combination. In total we have $109$ clusters and in each cluster, we have $10$ training samples. We perform data recombination to create $50$ examples for each cluster.
Our template validation selects the top $5$ templates and perform template refinement on these templates.
Our template refinement process uses $-2$ log probability as a threshold for removing ungeneralizable spans.

\subsection{In-domain Evaluation}
\label{sec:app:in-domain}
\paragraph{Additional Details for Experiment Setup.}
On E2E, the \texttt{familyFriendly} field is a binary field with values being either ``yes'' or ``no''.
To accomodate template-based generation, we replace ``yes'' with ``family friendly'' and ``family-friendly'' and replace ``no'' with ``not family friendly'' and ``not family-friendly''.
We augment E2E input $d$ with article words \texttt{[article: [a, an]]}.

On SynthBio, we augment inputs with values listed in 
\Cref{tab:synthbio_aug}. 
For \texttt{article}, \texttt{be}, and \texttt{number}, we include them as multiple value options in the input.
For pronouns and relation, we assign the correct value based on the \texttt{gender} field in the input.
We parse all dates into day, month, and year and create separate fields to support different data formats in the templates.

\begin{table*}[]
    \centering
    \begin{tabular}{c|c}
        \toprule
        Data Field & Data Value \\
        \midrule
        article &  a, an\\
        \midrule
        be & is, are, was, were \\
        \midrule
        number & \Centerstack{one, two, three, four, five,\\ six, seven, eight, nine, ten}\\
        \midrule
        pronoun\_a & he, she, they \\
        \midrule
        pronounce\_b & him, her, them \\
        \midrule
        pronounce\_c & his, her, their \\
        \midrule
        relation & son, daughter \\
        \bottomrule
    \end{tabular}
    \caption{Data fields and values we used for augmenting SynthBio input.}
    \label{tab:synthbio_aug}
\end{table*}

\paragraph{Implementation of Faithfulness Evaluation.}
\begin{table*}[ht]
    \centering
    \begin{tabular}{c|c|c}
    \toprule
        field & value & phrasing \\
    \midrule
food & Fast food & \Centerstack{Fast food \\ fast food}\\
\midrule
familyFriendly & yes & \Centerstack{is family friendly \\ is kid friendly \\ is children friendly \\ is family-friendly \\ is child friendly \\ is a family friendly \\ is a kid friendly \\ is a children friendly \\ is a family-friendly \\ is a child friendly \\ for a family friendly \\ for a kid friendly \\ for a children friendly \\ for a family-friendly \\ for a child friendly}\\
\midrule
familyFriendly & no & \Centerstack{not family friendly \\ not kid friendly \\ not children friendly \\ not family-friendly \\ not child friendly \\ non family-friendly \\ non-family-friendly \\ non family friendly \\ non-family friendly \\ non children friendly \\ non child friendly}\\
\midrule
customer rating & 1 out of 5 & \Centerstack{1 out of 5 \\ low customer rating \\ one star \\ 1 star}\\
\midrule
customer rating & 3 out of 5 & \Centerstack{3 out of 5 \\ customer rating is average \\ average customer rating \\ three star \\ moderate customer rating \\ 3 star}\\
\midrule
customer rating & 5 out of 5 & \Centerstack{5 out of 5 \\ high customer rating \\ five star \\ 5 star}\\
    \bottomrule
    \end{tabular}
    \caption{A collection of common paraphrases of given input data. We use this phrasing collection to perform a matching-based faithfulness evaluation for E2E. The second half of this table is in \Cref{tab:fact-2}.}
    \label{tab:fact-1}
\end{table*}

\begin{table*}[ht]
    \centering
    \begin{tabular}{c|c|c}
    \toprule
        field & value & phrasing \\
    \midrule
customer rating & high & \Centerstack{5 out of 5 \\ high customer rating \\ five star \\ 5 star}\\
\midrule
customer rating & average & \Centerstack{3 out of 5 \\ customer rating is average \\ average customer rating \\ three star \\ 3 star}\\
\midrule
customer rating & low & \Centerstack{1 out of 5 \\ low customer rating \\ one star \\ 1 star}\\
\midrule
priceRange & less than £20 & \Centerstack{less than £20 \\ cheap \\ low price range \\ low-priced \\ low priced}\\
\midrule
priceRange & £20-25 & \Centerstack{£20-25 \\ moderate price range \\ average price range \\ moderately priced \\ moderate prices \\ average priced}\\
\midrule
priceRange & more than £30 & \Centerstack{more than £30 \\ high price range \\ high priced \\ expensive \\ price range is high}\\
\midrule
priceRange & low & \Centerstack{low price range \\ low-priced}\\
\midrule
priceRange & cheap & \Centerstack{cheap \\ low price range \\ low priced}\\
\midrule
priceRange & moderate & \Centerstack{moderate price range \\ moderately priced \\ price range is moderate \\ moderate prices \\ average prices}\\
\midrule
priceRange & high & \Centerstack{high price range \\ high priced \\ expensive \\ price range is high}\\
    \bottomrule
    \end{tabular}
    \caption{A collection of common paraphrases of given input data. We use this phrasing collection to perform a matching-based faithfulness evaluation for E2E. The first half of this table is in \Cref{tab:fact-1}.}
    \label{tab:fact-2}
\end{table*}
We present the phrasing collection we used for matching output in \Cref{tab:fact-1} and \Cref{tab:fact-2}.
We use this phrasing collection to perform matching based faithfulness evaluation.
We consider a phrase in an output to have a precision error if it matches with a field and value pair that is not present in the input data.
We consider an output as having recall error $E_{\text{recall}}$ if we cannot identify any phrase in the output that corresponds to some field and value pair in the input data
Because our phrasing collection is imperfect and alternative phrasing may exist, we expect $E_{\text{precision}}$ to be an underestimate and $E_{\text{recall}}$ to be an overestimate of actual errors.

\paragraph{Additional Results for \Cref{sec:in-domain}.}
\begin{table*}
\small
\begin{center}
\begin{tabular}{c|c|cccccccccc}
\toprule
 Split & Methods & \bleu $\uparrow$ & \nist $\uparrow$ & \meteor $\uparrow$ & \rougel $\uparrow$ & \cider $\uparrow$ & $E_{\text{precision}} \downarrow$ & $E_{\text{recall}} \downarrow$ \\
\midrule
\multirow{4}{*}{Test} & BART & $\mathbf{66.2}\pm 0.5$ & $\mathbf{8.5}\pm 0.0$ & $\mathbf{43.1}\pm 0.2$ & $\mathbf{68.4}\pm 0.7$ & $\mathbf{2.2}\pm 0.0$ & $6.0\pm 2.9$ & $\mathbf{376.3}\pm 48.1$\\
\cmidrule{2-9}
& TempLM & $61.5\pm 1.0$ & $8.0\pm 0.1$ & $41.0\pm 0.8$ & $64.5\pm 0.8$ & $2.1\pm 0.1$ & $0.0\pm 0.0$ & $471.7\pm 62.9$\\
\cmidrule{2-9}
& $\text{NTemp}^{\dagger}$ & 55.17 & 7.14 & 41.91 & 65.70 & 1.70 & 7 & 539 \\
& TempClassic & $52.1\pm 2.0$ & $7.3\pm 0.1$ & $41.7\pm 1.0$ & $62.2\pm 2.3$ & $1.9\pm 0.1$ & $46.7\pm 25.4$ & $451.7\pm 36.9$\\
& SUB & $45.3\pm 1.9$ & $6.9\pm 0.2$ & $40.0\pm 0.2$ & $55.6\pm 2.4$ & $1.4\pm 0.1$ & $110.7\pm 36.2$ & $421.0\pm 12.7$\\
\midrule
\multirow{4}{*}{Valid.} &BART & $\mathbf{70.8}\pm 0.7$ & $\mathbf{8.3}\pm 0.1$ & $\mathbf{47.0}\pm 0.1$ & $\mathbf{72.8}\pm 0.2$ & $\mathbf{2.4}\pm 0.0$ & $5.0\pm 1.5$ & $\mathbf{182.0}\pm 11.8$\\
\cmidrule{2-9}
& TempLM & $64.8\pm 0.6$ & $8.0\pm 0.0$ & $43.1\pm 0.4$ & $67.8\pm 0.2$ & $2.2\pm 0.0$ & $\mathbf{0.0}\pm 0.0$ & $308.7\pm 4.3$\\
\cmidrule{2-9}
& $\text{NTemp}^{\dagger}$ & 64.53 & 7.66 & 42.46 & 68.60 & 1.82 & 7 & 539 \\
& TempClassic & $52.2\pm 0.6$ & $7.2\pm 0.0$ & $40.9\pm 0.2$ & $60.7\pm 0.9$ & $1.7\pm 0.0$ & $92.7\pm 6.1$ & $401.0\pm 13.2$\\
& SUB & $43.0\pm 0.4$ & $6.6\pm 0.1$ & $39.4\pm 0.2$ & $55.0\pm 0.4$ & $1.3\pm 0.0$ & $85.3\pm 16.9$ & $409.7\pm 13.7$\\
\bottomrule
\end{tabular}
    \caption{Evaluation of systems trained on the subsampled E2E datasets.}
    \label{tab:full_subsample}
\end{center}
\end{table*}

\begin{table*}
\begin{center}
\begin{tabular}{c|c|cccc}
\toprule
&  & \bleu & \bertf & \rougel \\
\midrule
\multirow{4}{*}{Test} & BART & $\mathbf{40.8}\pm 0.2$ & $\mathbf{55.2}\pm 0.1$ & $\mathbf{48.4}\pm 0.2$\\
\cmidrule{2-5}
& TempLM & $40.3\pm 0.3$ & $54.3\pm 0.1$ & $48.3\pm 0.1$\\
\cmidrule{2-5}
& TempClassic & $36.6\pm 0.2$ & $48.8\pm 0.1$ & $43.1\pm 0.1$\\
& SUB & $14.1\pm 0.1$ & $18.9\pm 0.1$ & $26.4\pm 0.1$ \\
\midrule
\multirow{4}{*}{Valid} & BART & $\mathbf{41.7}\pm 0.3$ & $\mathbf{55.6}\pm 0.1$ & $48.8\pm 0.1$ \\
\cmidrule{2-5}
& TempLM & $41.3\pm 0.2$ & $55.2\pm 0.2$ & $\mathbf{49.1}\pm 0.2$ \\
\cmidrule{2-5}
& TempClassic & $35.1\pm 0.2$ & $47.7\pm 0.1$ & $42.0\pm 0.1$ \\
& SUB & $14.0\pm 0.1$ & $19.0\pm 0.1$ & $26.4\pm 0.0$\\
\bottomrule
\end{tabular}
\end{center}
\caption{Automatic evaluation results on the SynthBio test and validation sets.}
\label{tab:full_synthbio}
\end{table*}

\begin{table*}
\scriptsize
\begin{center}
\begin{tabular}{c|c|ccccccccccc}
\toprule
 Split & Methods & \bleu $\uparrow$ & \nist $\uparrow$ & \meteor $\uparrow$ & \rougel $\uparrow$ & \cider $\uparrow$ & $E_{\text{precision}} \downarrow$ & $E_{\text{recall}} \downarrow$ & $\#$. Templates \\
\midrule
\multirow{4}{*}{Test} & BART & $\mathbf{67.1}\pm 0.2$ & $\mathbf{8.7}\pm 0.0$ & $\mathbf{45.2}\pm 0.0$ & $\mathbf{69.5}\pm 0.1$ & $\mathbf{2.3}\pm 0.0$ & $\mathbf{0.0}\pm 0.0$ & $\mathbf{110.7}\pm 5.2$ & N/A\\
\cmidrule{2-10}
& TempLM & $57.4\pm 0.6$ & $7.6\pm 0.0$ & $41.0\pm 0.3$ & $65.8\pm 0.3$ & $2.0\pm 0.0$ & $\mathbf{0.0}\pm 0.0$ & $506.7\pm 15.6$ & $\mathbf{509}$ \\
\cmidrule{2-10}
& $\text{NTemp}^{\dagger}$ & $55.17$ & $7.14$ & $41.91$ & $65.70$ & $1.70$ & $7$ & $539$ & N/A \\
& TempClassic & $58.2\pm 0.0$ & $7.5\pm 0.0$ & $43.7\pm 0.0$ & $67.6\pm 0.0$ & $2.2\pm 0.0$ & $\mathbf{0.0}\pm 0.0$ & $516.0\pm 1.0$ & $39964$\\
& SUB & $36.8\pm 0.2$ & $5.9\pm 0.0$ & $39.5\pm 0.1$ & $51.2\pm 0.2$ & $0.81\pm 1.6$ & $183.7\pm 3.2$ & $416.3\pm 1.5$ & $39964$\\
\midrule
\multirow{4}{*}{Valid.} &BART & $\mathbf{69.8}\pm 0.1$ & $\mathbf{8.4}\pm 0.0$ & $\mathbf{47.6}\pm 0.1$ & $\mathbf{74.3}\pm 0.1$ & $\mathbf{2.5}\pm 0.0$ & $0.3\pm 0.3$ & $\mathbf{256.3}\pm 5.8$ & N/A \\
\cmidrule{2-10}
& TempLM & $65.5\pm 0.1$ & $7.8\pm 0.0$ & $45.7\pm 0.1$ & $71.9\pm 0.2$ & $2.4\pm 0.0$ & $\mathbf{0.0}\pm 0.0$ & $365.7\pm 9.4$ & $\mathbf{509}$ \\
\cmidrule{2-10}
& $\text{NTemp}^{\dagger}$ & $64.53$ & $7.66$ & $42.46$ & $68.60$ & $1.82$ & $7$ & $539$ & N/A \\
& TempClassic & $64.6\pm 0.1$ & $7.8\pm 0.0$ & $46.0\pm 0.0$ & $71.3\pm 0.0$ & $2.4\pm 0.0$ & $4.0\pm 0.0$ & $425.7\pm 0.9$ & $39964$ \\
& SUB & $35.9\pm 0.1$ & $5.6\pm 0.0$ & $38.8\pm 0.1$ & $51.7\pm 0.1$ & $0.73\pm 0.4$ & $136.0\pm 3.8$ & $374.0\pm 1.7$ & $39964$\\
\bottomrule
\end{tabular}
    \caption{Evaluation of systems trained on the full E2E training set.}
    \label{tab:e2e_full_data}
\end{center}
\end{table*}

We present a full set of metrics scores for subsampled E2E and SynthBio in \Cref{tab:full_subsample} and \Cref{tab:full_synthbio}.
We make similar observations as in \Cref{sec:in-domain}: first, \templm{} is the most faithful system on E2E, never producing any precision error; second,  \templm{} is more fluent than other template systems, achieves better scores with the most of the metrics (BLEU, NIST, CIDEr), and on-par scores with METEOR and ROUGE-L.

We carry out the same experiment on E2E with models trained on the full dataset and present the results in \Cref{tab:e2e_full_data}.
We observe that similar to \templm{} is the only model that never produces unfaithful on both the test set and the validation set. 
BART becomes more faithful with more training data.
Similar to the experiments on the subsampled training set, \templm{} achieves better fluency than NTemp and SUB.
One different observation from \Cref{tab:e2e_full_data} is that TempClassic achieves much better fluency and faithfulness. 
This is because by leveraging the full training data, TempClassic obtains a large number of template ($39964$).
While using a large number of template is helpful, it makes PLM-based inference infeasibly slow, requiring hours of computation to perform inference on the test and validation sets.
Having many templates also makes the template set less interpretable by human inspectors.
Therefore, we consider TempClassic an impractical baseline.

\subsection{Out-of-domain Evaluation}
\label{sec:app:ood}
\Cref{tab:ood_entities} displays the list of entities we used for creating the $54$ OOD examples we used in our evaluation.
\Cref{tab:ood_example} shows example outputs from the BART model finetuned on the downsampled E2E data with OOD input.
We find that BART often confuses the entity in the \texttt{area} field with \texttt{name} or ignores the input value and hallucinate ``city centre.''

\begin{table*}[h!]
    \centering
    \begin{tabular}{c|c}
    \toprule
        Area  & Times Square, Central Park, Union Station \\ 
        \midrule
        Food  & Thai, Russian, German \\ 
        \midrule
        Restaurant  & McDonald's, Subway, Starbucks, KFC \\ 
        \midrule
        Near  & McDonald's, Subway, Starbucks, KFC \\ 
    \bottomrule
    \end{tabular}
    \caption{List of novel entities used for creating OOD examples.}
    \label{tab:ood_entities}
\end{table*}

\begin{table*}[h!]
    \centering
    \begin{tabular}{c|c}
    \toprule
        Input data  & \Centerstack{\{area: Times Square, eatType: restaurant, food: Thai,\\ name: McDonald's, near: Subway\}} \\ 
        \midrule
        BART output & Near Subway in {\color{red}the city center} is a restaurant that serves Thai food called the Mcdonald's. \\
        \midrule
        Input data  & \Centerstack{\{area: Central Park, eatType: restaurant, food: Thai,\\ name: McDonald's, near: Subway\}} \\ 
        \midrule
        BART output &  {\color{red} Central park} is a restaurant that serves thai food. it is located near subway.\\
        \midrule
        Input data  & \Centerstack{\{area: Times Square, eatType: restaurant, food: Russian, \\name: McDonald's, near: KFC\}} \\ 
        \midrule
        BART output & Near KFC in the city center is a restaurant called {\color{red} the Times Square}. it serves Russian food. \\
        \midrule
        Input data  & \Centerstack{\{area: Union Station, eatType: restaurant, food: German,\\ name: Subway, near: Starbucks\}} \\ 
        \midrule
        BART output & In {\color{red}the center of the city} near Starbucks is a restaurant called Subway. \\
    \bottomrule
    \end{tabular}
    \caption{Examples OOD output from the BART model finetuned on the downsampled E2E dataset. We color unfaithful information {\color{red} red}.}
    \label{tab:ood_example}
\end{table*}

\subsection{Human Study}
\label{sec:app:human}
We present a full list of metric scores that we used to evaluate our human study in \Cref{tab:human-study-full}.
We have similar observations as in \Cref{sec:human} that \templm{} extracts more fluent templates than our template writers.
We append our instruction for template writers and screenshots of our interface to the end of this document.

\begin{table*}[]
    \begin{center}
    \begin{tabular}{c|l|ccccc}
    \toprule
    &  & \bleu & \bertf & \rougeone & \rougetwo & \rougel \\
    \midrule
    \multirow{5}{*}{\Centerstack{Writer \\Cluster}} & Human & $37.3\pm 1.5$ & $51.3\pm 2.3$ & $64.5\pm 1.1$ & $41.1\pm 1.6$ & $44.9\pm 1.7$\\
     & \shortstack[l]{Human \\Ensemble} & $39.1$ & $54.0$ & $63.7$ & $44.1$ & $47.3$\\
     & BART & $44.0\pm 0.2$ & $58.5\pm 0.2$ & $\mathbf{70.6}\pm 0.3$ & $45.8\pm 0.3$ & $50.9\pm 0.2$\\
     & TempLM & $\mathbf{44.3}\pm 1.3$ & $\mathbf{58.8}\pm 1.0$ & $68.6\pm 1.1$ & $\mathbf{46.8}\pm 1.3$ & $\mathbf{51.8}\pm 0.7$\\
    \midrule
    \multirow{5}{*}{\Centerstack{Spy \\Cluster}} & Human & $24.9\pm 2.0$ & $42.2\pm 4.4$ & $54.8\pm 2.0$ & $34.8\pm 0.6$ & $40.5\pm 1.2$\\
     & \shortstack[l]{Human \\Ensemble} & $32.1$ & $48.5$ & $57.2$ & $37.2$ & $40.7$\\
     & BART & $\mathbf{40.5}\pm 0.4$ & $\mathbf{55.4}\pm 0.1$ & $\mathbf{68.2}\pm 0.4$ & $\mathbf{42.7}\pm 0.3$ & $\mathbf{46.5}\pm 0.1$\\
     & TempLM & $34.4\pm 2.4$ & $50.8\pm 0.9$ & $61.4\pm 0.9$ & $39.8\pm 1.2$ & $44.1\pm 0.4$\\
    \bottomrule
    \end{tabular}
    \end{center}
    \caption{Human study results on two clusters of the SynthBio test set. We observe that human written templates cannot achieve high metric scores even in the ensemble setting, showcasing the difficulty of writing templates and the efficacy of TempLM.}
    \label{tab:human-study-full}
\end{table*}

\clearpage


\section*{Supplementary material}

# Designing templates for data to text conversion

**Goal:** <mark>Write (ideally ten or more) templates</mark> that generate realistic biography
**Time:** 30 minutes

## 1. What is this task?

Your goal is to write a set of *templates* that can be used to automatically convert data into text. For example, consider this *data* which have three field and value pairs:

| Field | Value |
|---|---|
| name | Ramazan Inal |
| nationality | Turkish |
| occupation | writer |

In order to automatically generate this *text* from the data:

<p align="center">Ramazan Inal is a Turkish writer.</p>

we can create this template:

<p align="center">[name] is a [nationality] [occupation].</p>

and our system will deterministically replace each field with the value specified in the data.

<p align="center">
[name] → Ramazan Inal<br>
[nationality] → Turkish<br>
[occupation] → writer
</p>

<p align="center">[name] is a [nationality] [occupation]. → Ramazan Inal is a Turkish writer.</p>

Because we want to make templates *flexible* so that they can account for potential grammatical changes necessary for different values (e.g. "<u>a</u> Turkish writer" vs. "<u>an</u> English writer"), we added these additional fields and possible values to all input data:

| Field | Value |
|---|---|

| | |
|---|---|
| be | One of the following: is, are, was, were |
| article | One of the following: a, an |
| number | One of the following: One, two, three, four, five, six, seven, eight, nine, ten |

Therefore, the final template with these additional fields and values will be:

[name] [be] [article] [nationality] [occupation].

[name] → Ramazan Inal
[be] → is
[article] → a
[nationality] → Turkish
[occupation] → writer

[name] [be] [article] [nationality] [occupation]. → Ramazan Inal is a Turkish writer.

Note that sometimes, *not all fields are used* to generate the text. In the previous example, the number field is not used anywhere in the text, hence no need to be specified in the template.

## 2. What is the goal?

Given hundreds of pairs of such data and desired texts, your goal is to **write ten or more templates** that *can best represent the given data and text pairs* as well as *can be used to generate realistic biography for new data*.

For example, the previous template can be used with new data to generate biography as follows:

Template:

[name] [be] [article] [nationality] [occupation].

New data:

| Field | Value |
|---|---|
| name | Joseph Duch |

| | |
|---|---|
| gender | non-binary |
| nationality | Andorran |
| occupation | writer |
| be | One of the following: is, are, was, were |
| article | One of the following: a, an |
| number | One of the following: One, two, three, four, five, six, seven, eight, nine, ten |

Automatically generated text:

Joseph Duch is a Andorran writer.

### 3. How do I do this task?

1. Click one of the links to start: [writer][spy]
    a. Please do not refresh your window! The timer will be reset and you will start over.
    b. We suggest that you maximize the window and zoom out so that you can browse the data easily.

2. In the left panel, you will see all pairs of data and desired texts. **Click through them and** *get yourself familiar with fields, values, and texts.*
3. In the right panel, you will see a **text editor** where you can write templates while browsing

multiple data and desired texts at the same time. Please enclose the field names with brackets (e.g. [name]). Valid field names will be colored in **orange**.

a. Each time you write a template, click the "add a template" button in the right panel, copy and paste your template, and click the "save" button.

b. You can view the list of templates you have written by clicking the "list of templates" button in the left panel.

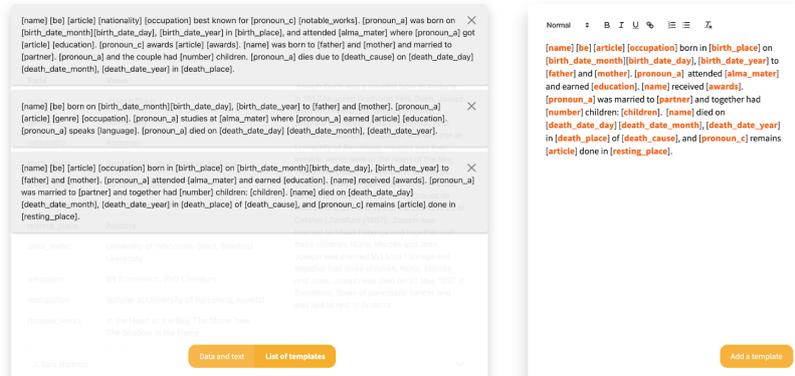

c. If necessary, you can delete templates by clicking the close button next to each template in the list.
4. On the bottom of the screen, you will see a counter for the number of templates and a timer.
5. When you are done, click the finish button next to the timer to save your templates. Share the verification code you got with Mina and share the templates you wrote with Tianyi.



\end{document}